\documentclass{IEEE_lsens}
\usepackage{textcomp}

\usepackage[noadjust]{cite}

\ifCLASSINFOpdf
\else
\fi
\usepackage[T1]{fontenc} 
\usepackage{amsmath}
\usepackage[cmintegrals]{newtxmath}
\usepackage{bm}

\usepackage[ruled,vlined]{algorithm2e}
\usepackage{graphicx}

\usepackage{array}
\usepackage{url}
\ifCLASSINFOpdf
\else
\fi
\providecommand{\hypersetup}[1]{\relax}

\hypersetup{pdftitle={Bare Demo of IEEE\_lsens.cls for IEEE Sensors Letters},
pdfsubject={Typesetting},
pdfauthor={Michael D. Shell},
pdfkeywords={Class, IEEE, IEEE\_lsens, IEEE Sensors Letters, LaTeX, Typesetting, TeX}}

\begin{document}

\markboth{Vol.~X, No.~X, XX~XXXX}{XXXXXXX}

\IEEELSENSarticlesubject{Sensor Applications}

%
\title{Deep Convolutional Generative Adversarial Network Based Food Recognition Using Partially Labeled Data}

%
\author{\IEEEauthorblockN{Bappaditya Mandal\IEEEauthorrefmark{1}, N. B. Puhan\IEEEauthorrefmark{2} and Avijit Verma\IEEEauthorrefmark{2}}
\IEEEauthorblockA{\IEEEauthorrefmark{1}School of Computing and Mathematics, Keele University, Staffordshire ST5 5BG, United Kingdom.\\
\IEEEauthorrefmark{2}School of Electrical Sciences, Indian Institute of Technology, Bhubaneswar, Odisha 751013, India.}%
\thanks{Corresponding author: B. Mandal (e-mail: b.mandal@keele.ac.uk).}%
}
%
%
%

\IEEELSENSmanuscriptreceived{Manuscript received XX XX, XXXX}

\IEEEtitleabstractindextext{%
\begin{abstract}
Traditional machine learning algorithms using hand-crafted feature extraction techniques (such as local binary pattern) have limited accuracy because of high variation in images of the same class (or intra-class variation) for food recognition task. In recent works, convolutional neural networks (CNN) have been applied to this task with better results than all previously reported methods. However, they perform best when trained with large amount of annotated (labeled) food images. This is problematic when obtained in large volume, because they are expensive, laborious and impractical. Our work aims at developing an efficient deep CNN learning-based method for food recognition alleviating these limitations by using partially labeled training data on generative adversarial networks (GANs). We make new enhancements to the unsupervised training architecture introduced by Goodfellow \emph{et al.} (2014), which was originally aimed at generating new data by sampling a dataset. In this work, we make modifications to deep convolutional GANs to make them robust and efficient for classifying food images. Experimental results on benchmarking datasets show the superiority of our proposed method as compared to the current-state-of-the-art methodologies even when trained with partially labeled training data.
\end{abstract}

\begin{IEEEkeywords}
Generative Adversarial Network, Deep CNN, Semi-Supervised Learning, Food Recognition.
\end{IEEEkeywords}}


\maketitle

\section{Introduction and Related Work}

A substantial increase has been witnessed in the health consciousness amongst masses due to increasing health risks. Diabetes, obesity and cholesterol cases are being increasingly reported every year. World Health Organisation (WHO) has reported that the global prevalence of diabetes has doubled over 1980 to 2014 \cite{WHO2016}. Improper diet and excessive calorie intake have been major factors causing these health risks and hence keeping a check on the calorie consumption can help avoid this risk \cite{Farooq1, Sazonov1}. With the advent of smart devices, apps for multi-modal data logging mechanism, which can be used anywhere and anytime are in high demand. At present, the existing mobile apps like MyFitnessPal \cite{MyFitnessPal}, LoseIt \cite{FitNow} and others work on manual data entry. This is cumbersome for most people hindering the long-term usability of such apps. Mobile cameras are used to capture image/video data and then typically rely on expert nutritionists to analyze the image offline (at a particular time) \cite{Schoeller1}. Crowd sourcing has also been used as an approach to analyze the food images. However, both these methods are costly, inefficient and slow; thus, they are not suitable for widespread usage.

Appearance of any food is highly characterized by the recipe, ingredients used, style of preparation and others. They exhibit large intra-class variation in terms of size, color, shape, texture, viewpoint and others \cite{Zhang4, Bi1}. In this case, classic feature descriptors like speeded up robust features (SURF), histogram of gradient (HOG), spatial pyramidal pooling, bag of scale invariant feature transform (SIFT), color correlogram, etc. can only succeed when used for laboratory generated small datasets. Even popular subspace learning algorithms \cite{Mandal12, Mandal6, Mandal14} follow the similar trend. Typically, a SVM is trained using these extracted features or a combination of these features \cite{Martinel1}. To improve accuracy for this task, researchers have worked towards estimating the region of the image in which the food item is present. One way to do this is to use standard segmentation and object detection methods or ask the user to input a bounding box giving this information as done in \cite{Matsuda1}. Another approach is to use convolutional neural network (CNN) based semantic segmentation, which has shown an increase in accuracy \cite{Yanai1}. Some approaches use ingredient-level features to recognize the food items \cite{Chen5}. Pairwise statistics of local features have also been applied for food recognition task \cite{Shulin1}.

Deep CNN learning algorithms overcome the drawbacks of traditional machine learning algorithms based on hand designed features. These algorithms have the inherent capability to mimic human information processing systems and work very well with images \cite{Krizhevsky1}, in many applications \cite{Liu14} including food recognition \cite{Pandey1}. In the recent years, CNN based food recognition has shown excellent results even on extensive and large datasets with non-uniform background \cite{Yanai1}. Recent reviews on recurrent neural networks (RNN) \cite{Salehinejad1}, shows the similar trend for learning long-term dependencies in sequential images (video) and time-series data. A common problem to all these existing methodologies is that they would need large amount of labeled training data to perform reasonably well, which is difficult and expensive to obtain when required in large amount.

\subsection{Related Generative Adversarial Network Architectures}
One of the attractive features in generative adversarial network (GAN) based modeling is that it does not require labeled data. Its learning approach is classified into generative and discriminative models: A \textit{generative} model is trained to learn the joint probability of the input data and output class labels simultaneously, i.e. $P(x,y)$. This can be used to infer the conditional distribution $P(y|x)$ by applying Bayes rule. More importantly, the joint probability learned can be used for other purposes, such as generating new $(x, y)$ samples. A \textit{discriminative} model is trained to learn a target function that maps the input data $x$ to a set of output class labels $y$. Mathematically, it approximates the conditional distribution $P(y|x)$. While both types of models: generative and discriminative have their use cases, generative models have the ability to model the internal nature of the input data even in the absence of any labels. In our current generation real world applications, unlabelled data is abundant and easy to obtain. The cost of acquiring labeled data can sometimes be too high to justify. In such cases, generative models provide a desirable solution.

As summarized in Algorithm \ref{GAN}, the principle idea behind a GAN is to pit a discriminative framework against a generative one. Thus, the two component neural networks in a GAN, discriminative and generative, act as adversaries. The generative network is given Gaussian noise as input and is trained to generate samples indistinguishable from the real samples. The discriminator network is given both the generated fake samples and corresponding real samples and is trained to identify the fake sample. In the introductory GAN work, Goodfellow \emph{et al.} \cite{Goodfellow1} used fully connected neural networks for both the generator and discriminator. Consequently, this architecture was applied to standard image datasets for testing purposes, specifically MNIST (handwritten digits) \cite{LeCun1} and CIFAR-10 (natural images) \cite{Krizhevsky2} datasets.

It is evident from the literature and benchmarking competitions that convolutional neural networks (CNNs) are extremely well suited to image data \cite{Krizhevsky1} analysis, like our case for food recognition \cite{Pandey1}. Initial experiments conducted on small datasets, like CIFAR-10, showed that achieving convergence in convolutional GANs  was harder than in CNNs, with similar computational power as that used for supervised learning. One solution to this problem was to use a cascade of CNNs with a Laplacian pyramid framework presented by Denton \emph{et al.} (LAPGAN) \cite{Denton1}. This essentially decomposed the generation process with the images generated in a coarse to fine manner. A group of network architectures called DCGAN (deep convolutional GAN) proposed by Radford \emph{et al.} \cite{Radford1}, showed promising results on image datasets. Here, a pair of convolutional discriminator and generator networks are simultaneously trained using strided and fractionally-strided convolutions. Thus, their aim was to learn the mapping from image space to the discriminator output space (down-sampling) and also to learn the mapping from a lower dimensional latent space to the image space (up-sampling). The widespread adoption of this group of GAN architectures for a number of applications makes it a natural choice for our semi-supervised task at hand.

\begin{algorithm} \label{GAN}
\DontPrintSemicolon
\SetKwInOut{Input}{Input}\SetKwInOut{Output}{output}
\Input{$I$ $\longleftarrow$\ Number of training iterations}
\Begin{
\For{number of training iterations}{
\For{$k$ steps}{
Sample mini batch of $m$ Gaussian noise samples $(z^1, z^2, ...., z^m)$ from noise prior $p_g(z)$.\;
Sample mini batch of $m$ examples $(x^1, x^2, ...., x^m)$ from data generating distribution $p_{data}(x)$.\;
Update the discriminator by ascending its stochastic gradient.
\begin{equation}
\nabla_{\theta_d} \frac{1}{m} \sum_{i=1}^{m} [ log D(x^i) + log (1-D(G(z^i))) ]
\end{equation}
}
Sample mini batch of $m$ Gaussian noise samples $(z^1, z^2, ...., z^m)$ from noise prior $p_g(z)$.\;
Update the generator by descending its stochastic gradient.
\begin{equation}
\nabla_{\theta_g} \frac{1}{m} \sum_{i=1}^{m} [ log (1-D(G(z^i))) ]
\end{equation}
}
}
\caption{Generative adversarial network training.}
\end{algorithm}

\begin{figure*}[htb]
\centering
\includegraphics[width=0.95\linewidth]{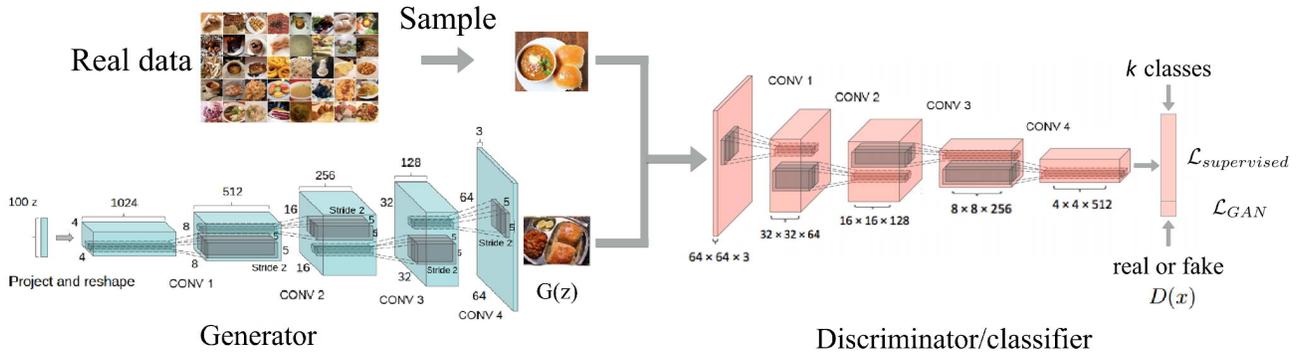} 
\caption{Block diagram for semi-supervised generative adversarial network for food recognition.}
\label{SSGAN} 
\end{figure*}

\section{Proposed Solution using GANs}
Our present focus is on the food recognition system that can deal with non-uniformity of the background and is extendable to out of sample test images. This would ensure that the system is robust and easy to use, the architecture has been designed for it using a semi-supervised approach as shown in Fig 1. The architecture has two parts: the Generator and the Discriminator. Although both are useful, discriminator is the network which learns the nature of the problem better and thus can be used as a multi-label classifier to recognize different food items. Below we describe our newly proposed semi-supervised GAN (SSGAN) architecture in detail.

\subsection{Semi-Supervised Generative Adversarial Networks}
For a general classifier, the input is the data point $x$ and the output is a $k$-dimensional vector of logits (inverse of the sigmoidal ``logistic'' function): $\varphi_{1}, \varphi_{2}, \ldots , \varphi_{k}$, which can be used to calculate class probabilities using softmax function: $p_{model}(y = i|x) = \frac{exp(\varphi_{i})}{\sum_{j=1}^{k} exp{(\varphi_{j})}}$, where $k$ is the number of classes. To train the model, we minimize the negative log-likelihood between $p_{model}(y|x)$ and the observed labels $y$. We add the fake generated samples from the generator $G$ to our dataset and also add another output class to our classifier to detect these fake samples. Now that we have $k+1$ classes, $p_{model}(y = k + 1 | x)$ determines the probability of the input data point being fake. This formulation also provides us the capability of learning from unlabeled data, provided it can be classified under one of the $k$ classes of real data by minimizing $[-log\ p_{model}(y \epsilon (1, \ldots , k)|x)]$.

In case of even distribution of data points, the loss function can be decomposed into two components, $\theta$, which is the negative log-likelihood of the label (supervised), when the data is real and $\delta$, which is the GAN loss (unsupervised).
\begin{equation} \label{eq:5.1}
\begin{aligned}
    L = -\mathbb{E}_{x,y \sim p_{data}(x,y)}[ log\ p_{model}(y|x) ] \\
    - \mathbb{E}_{x \sim G}[ log\ p_{model}(y = k+1 | x ) ].
\end{aligned}
\end{equation}
Total loss,
$L = \theta + \delta$, where
\begin{equation}\label{eq:5.2}
    \theta = -\mathbb{E}_{x,y \sim p_{data}(x,y)}[ log\ p_{model}(y|x, y < k+1) ]
\end{equation}
and
\begin{equation}
\begin{aligned}
    \delta = -\mathbb{E}_{x \sim p_{data}(x)}[ 1 - log\ p_{model}(y = k+1|x) ]\\ -
    \mathbb{E}_{x \sim G}[ log\ p_{model}(y = k+1 | x ) ].
\end{aligned}
\end{equation}

The outputs from the GAN discriminator network $D$ is an estimated probability that the input image is obtained from the data generating distribution. Traditionally, this is implemented with a feed-forward network ending in a single sigmoid unit, but in this work a softmax output layer is used as one unit for each of the $real$ classes and one $fake$ unit for generated images. It can be seen that $D$ has $k+1$ output units corresponding to $[class-1, class-2, . . ., class-k, fake]$. In this case, $D$ can also act as a classifier $C$. We use higher granularity labels for the half of the mini-batch that has been drawn from the data generating distribution. Training is performed on the discriminator/classifier network to minimize the negative log-likelihood corresponding to the given labels.

The generator $G$ is trained to maximize this negative log-likelihood. Our proposed SSGAN framework is summarized in Algorithm \ref{algo2}.
\begin{algorithm} \label{algo2}
\DontPrintSemicolon
\SetKwInOut{Input}{Input}\SetKwInOut{Output}{output}
\Input{$I$ $\longleftarrow$\ Number of training iterations}
\Begin{
\For{i = 1 \textbf{to} $I$}{
 Sample mini batch of $m$ Gaussian noise samples $(z^1, z^2, ...., z^m)$ from noise prior $p_g(z)$.\;

 Sample mini batch of $m$ examples $(x^1, x^2, ...., x^m)$ from data generating distribution $p_{data}(x)$.\;

 Perform gradient descent on the parameters of $D$ by calculating the gradient as:
 \begin{equation}
\nabla_{\theta_d} \frac{1}{2m} \sum_{i=1}^{2m} L \ (loss\ in\ equation\ \ref{eq:5.1})
\end{equation}

 Sample mini batch of $m$ Gaussian noise samples $(z^1, z^2, ...., z^m)$ from noise prior $p_g(z)$.\;

 Perform gradient descent on the parameters of $G$ by calculating the gradient as:
\begin{equation}
\nabla_{\theta_d} \frac{1}{m} \sum_{i=1}^{m}[ 1 - log\ p_{model}(y = K+1|x) ]
\end{equation}
}
}
\caption{Semi-supervised GAN (SSGAN) training algorithm.}
\end{algorithm}

\section{Experimental Results and Discussions}
We have conducted experiments on two datasets, ETH Food-101 \cite{Bossard1} and Indian Food Dataset \cite{Pandey1}. The experiments are performed on an NVIDIA Quadro P4000 GPU with 8GB VRAM. Our model details are: GAN type: DCGAN, Optimizer: ADAM, Activation: Leaky ReLu (in all hidden layers), Sigmoid (in discriminator's output layer). The model was run on test data simultaneously after every 100 training epochs. To address the instability of GANs while training, we have implemented feature matching by specifying a new objective for the generator side that prevents it from over training on the discriminator side, similar to as mentioned in \cite{SalimansGZCRC16}. Additionally, we implemented one-sided label smoothing and batch normalization for stable convergence as discussed in Salimans \emph{et al.} \cite{SalimansGZCRC16}. In our case, the stochastic layers are zero-centered Gaussian noise, with standard deviation of 0.5 for input and 0.5 for outputs of hidden layers. During the training process, for ETH Food-101 and Indian Food datasets, we used 10\% and 50\% data of each of the classes as unlabeled data, respectively. For comparisons, fine-tuned deep learned models are used from the popular AlexNet \cite{Krizhevsky1}, GoogLeNet \cite{Christian1}, residual network (ResNet) \cite{He4}, for all the methods, procedures are used as reported in \cite{Pandey1}. We also compared the new method (SSGAN) with the recently proposed Lukas \emph{et al.} \cite{Bossard1}, Kawano \emph{et al.} \cite{Kawano1}, Martinel \emph{et al.} \cite{Martinel1} and the ensemble of networks (Ensemble Net) in \cite{Pandey1}.

\subsection{Results on ETH Food-101 Dataset}
\begin{figure}
\centering
\begin{minipage}[b]{4.3cm}
\includegraphics*[height=3.4cm]{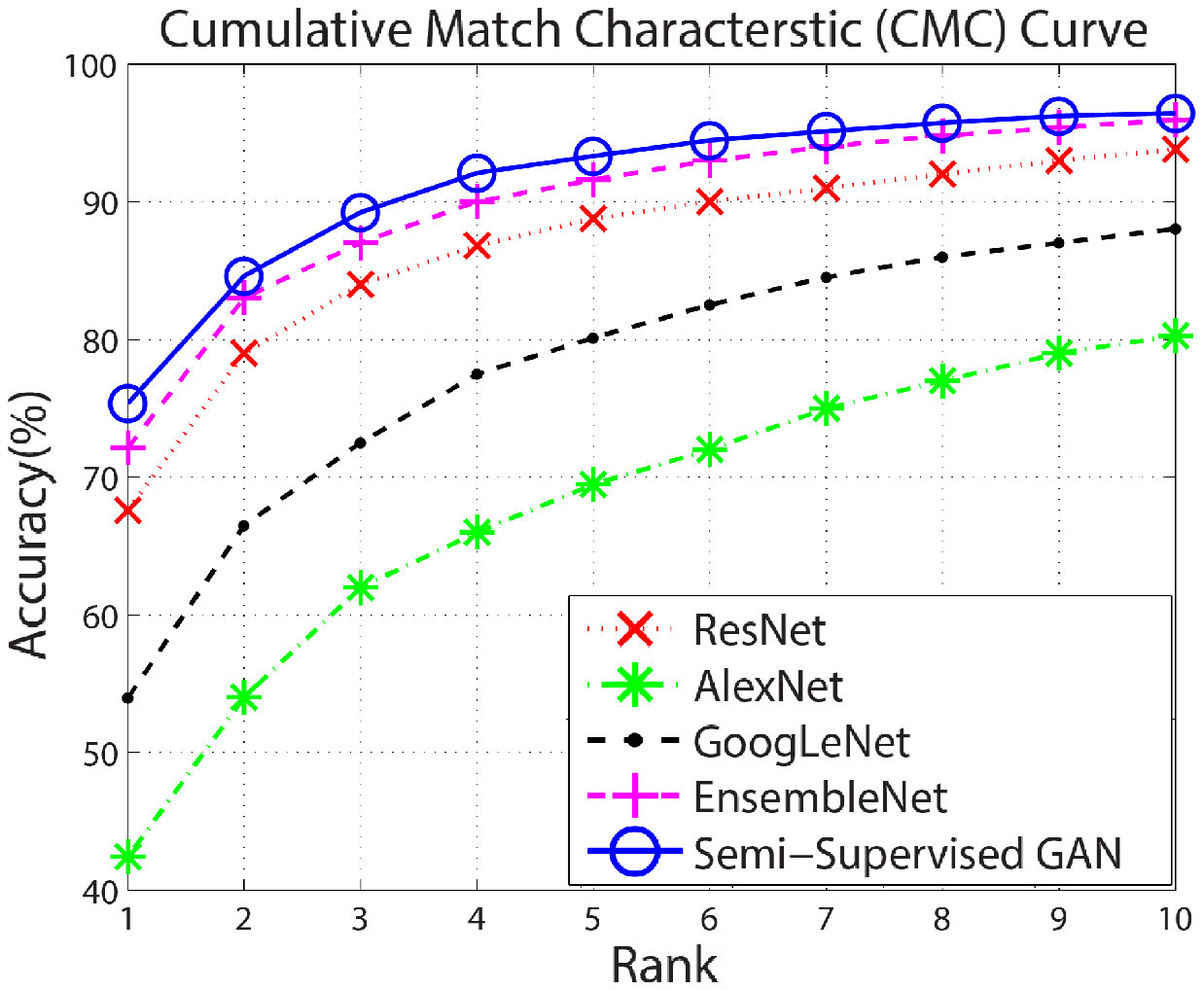}
\centering (a)
\end{minipage}
\begin{minipage}[b]{4.3cm}
\includegraphics*[height=3.4cm]{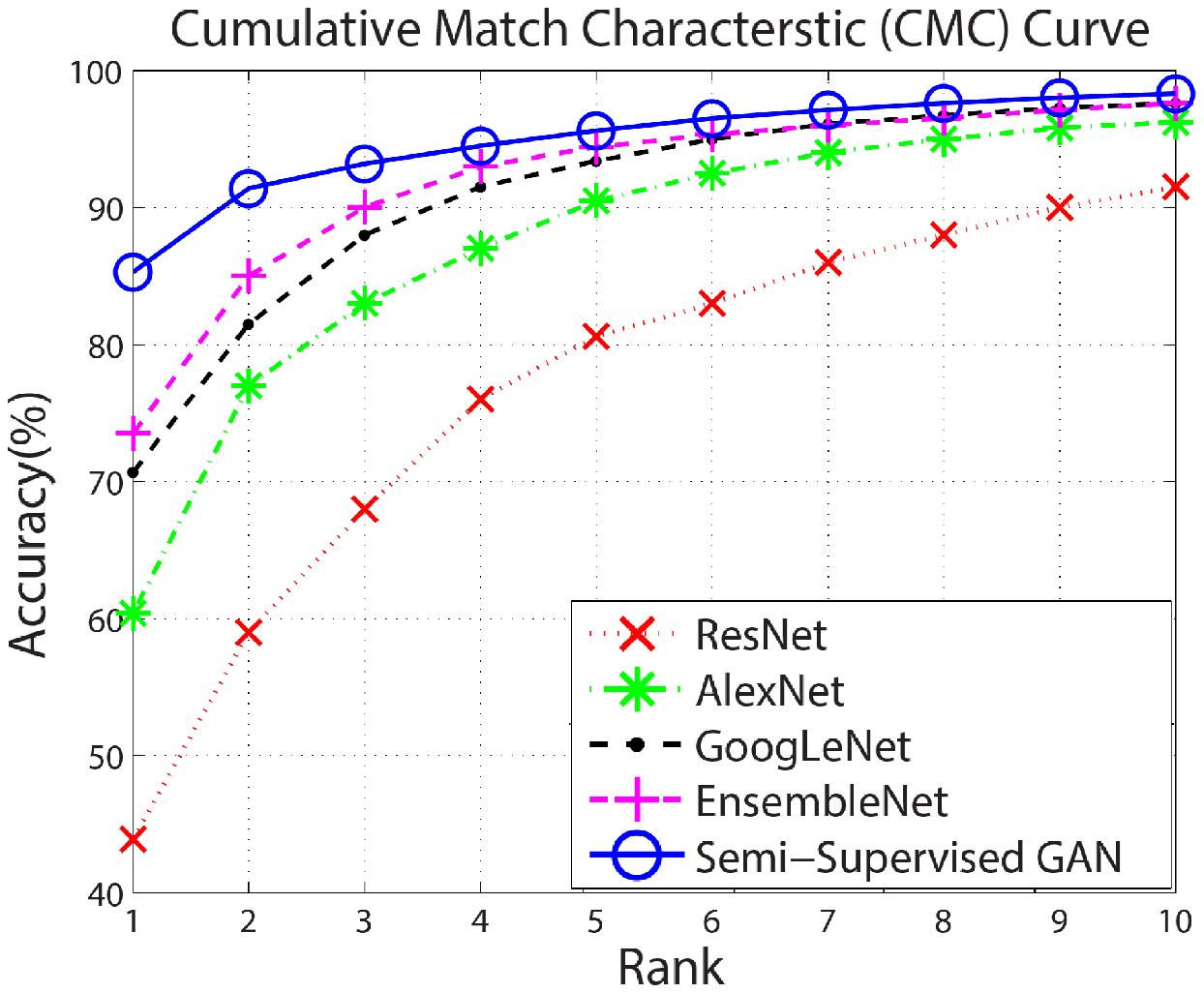}
\centering (b)
\end{minipage}
\caption{Accuracy vs rank plots using various existing CNN frameworks and SSGAN, (a) for ETH Food 101 Database and (b) for Indian Food Database.}
\label{GraphFood}
\end{figure}

ETH Food-101 \cite{Bossard1} is a database consisting of 1000 images per food class of 101 classes of most popular food categories picked up from foodspotting.com. The top 101 most popular and consistently named dishes were chosen and 750 training images were extracted. An addition of 250 test images were collected per class. While the collected test images were manually cleaned, the training images were not cleaned deliberately to retain some amount of noise. The idea is to increase the robustness of any classifier trained on the dataset. For our experiments, we follow the same training and testing protocols as discussed in \cite{Bossard1, Martinel1}. Fig. \ref{GraphFood} (a) shows the accuracy vs ranks plots up to top 10 ranks, where the rank $r : r \epsilon \{ 1, 2,..., 10 \}$ denotes the probability of retrieving at least one correct image among the top $r$ retrieved images. This cumulative matching curve (CMC) shows the overall performance of the proposed approach as the number of retrieved images changes. Table \ref{Table1} shows the Top-1, Top-5 and Top-10 accuracies using current state-of-the-art methodologies on this dataset.
\begin{table}[!htb] \tiny
    \begin{minipage}{.5\linewidth}
      \centering \addtolength{\tabcolsep}{-1pt}
        \caption{Accuracy (\%) for ETH Food-101 \& comparison with other methods.} \label{Table1}
        \begin{tabular}{|c|c|c|c|}
        \hline
        Network/Features & Top-1 & Top-5 & Top-10 \\
        \hline
        AlexNet & 42.42 & 69.46 & 80.26 \\
        \hline
        GoogLeNet & 53.96 & 80.11 & 88.04 \\
        \hline
        Lukas \emph{et al.} \cite{Bossard1} & 50.76 & - & - \\
        \hline
        Kawano \emph{et al.} \cite{Kawano1} & 53.50 & 81.60 & 89.70 \\
        \hline
        Martinel \emph{et al.} \cite{Martinel1} & 55.89 & 80.25 & 89.10 \\
        \hline
        ResNet \cite{He4} & 67.59 & 88.76 & 93.79 \\
        \hline
        Ensemble Net \cite{Pandey1} & 72.12 & 91.61 & 95.95 \\
        \hline
        \textbf{SSGAN} & \textbf{75.34} & \textbf{93.31} & \textbf{96.43} \\
        \hline
        \end{tabular}
    \end{minipage}
    \begin{minipage}{.5\linewidth}
    \centering \addtolength{\tabcolsep}{-1pt}
      \caption{Accuracy (\%) for Indian Food Database \& comparison with other methods.} \label{Table2}
      \begin{tabular}{|c|c|c|c|}
      \hline
      Network/Features & Top-1 & Top-5 & Top-10 \\
      \hline
      AlexNet & 60.40 & 90.50 & 96.20 \\
      \hline
      GoogLeNet & 70.70 & 93.40 & 97.60 \\
      \hline
      ResNet \cite{He4} & 43.90 & 80.60 & 91.50 \\
      \hline
      Ensemble Net \cite{Pandey1} & 73.50  & 94.40 & 97.60 \\
      \hline
      \textbf{SSGAN} & \textbf{85.30}  & \textbf{95.60} & \textbf{98.30} \\
      \hline
      \end{tabular}
    \end{minipage}
\end{table}

From Fig. \ref{GraphFood}(a) and Table \ref{Table1}, it is evident that our proposed semi-supervised GAN (SSGAN) has outperformed consistently all the current state-of-the-art methodologies on this large real-world food dataset.

\subsection{Results on Indian Food Dataset}
Indian food dataset is newly introduced in \cite{Pandey1} as an extensive collection of Indian food images comprising of 50 food classes with 100 images in each class. The classes were selected keeping in mind the extensive nature of Indian cuisine. Indian food differs in terms of shape, size, color and texture and is devoid of any generalized make-up. Because of the varied nature of the classes present in the dataset, it offers the best option to test a protocol and classifier for its robustness and accuracy. It consists of images from online sources like Food Spotting, Google in addition to images captured using hand-held mobile devices. Similar to the ETH Food-101 database protocol and same protocol as that in \cite{Pandey1}, we have set aside 80 food images per class for training and rest of the images are used for testing. Fig. \ref{GraphFood} (b) shows accuracy vs ranks plot up to top 10 ranks and Table \ref{Table2} shows the Top-1, Top-5 and Top-10 accuracies using the current state-of-the-art methodologies on this dataset. Both of these show that our proposed SSGAN model performs better at recognizing food images in comparison to other CNN frameworks for this dataset. Overall, it is evident from these Figures \ref{GraphFood} (a) \&  (b) and Tables \ref{Table1} \& \ref{Table2} that our proposed approach (SSGAN) outperforms the other methods consistently for all ranks on both these datasets.

\section{Conclusion and Future Work}
Food recognition is a very challenging task due to the presence of high intra-class variation in food appearance. This is due to the differences in method of preparation, ingredients used, various shapes, viewpoints and other factors. A classifier working well on one kind of cuisine may not give as good results on another type of cuisine. In this work, we have proposed a semi-supervised GANs based on deep convolutional neural network architecture approach to alleviate the shortcomings posed by lack of labeled images and also the classical image recognition problems in food datasets. We have performed experiments on the largest real-world food images ETH Food-101 dataset and the Indian Food dataset with partially labeled data. Experimental results show that the generative semi-supervised deep CNN approach proposed in this work outperforms the current state-of-the-art methodologies consistently for all the ranks for both the datasets even with partially labeled data. While GANs have the potential to improve the food recognition accuracy with partially labeled data, it is difficult to achieve stability and convergence during training. In future, we would try to improve the recognition accuracy with better and robust GAN architecture that could further reduce the usage of labeled training data.

\normalsize

%
%
\bibliographystyle{IEEEtran}
\bibliography{BiblioJan2018}

\begin{thebibliography}{10}
\providecommand{\url}[1]{#1}
\csname url@samestyle\endcsname
\providecommand{\newblock}{\relax}
\providecommand{\bibinfo}[2]{#2}
\providecommand{\BIBentrySTDinterwordspacing}{\spaceskip=0pt\relax}
\providecommand{\BIBentryALTinterwordstretchfactor}{4}
\providecommand{\BIBentryALTinterwordspacing}{\spaceskip=\fontdimen2\font plus
\BIBentryALTinterwordstretchfactor\fontdimen3\font minus
  \fontdimen4\font\relax}
\providecommand{\BIBforeignlanguage}[2]{{%
\expandafter\ifx\csname l@#1\endcsname\relax
\typeout{** WARNING: IEEEtran.bst: No hyphenation pattern has been}%
\typeout{** loaded for the language `#1'. Using the pattern for}%
\typeout{** the default language instead.}%
\else
\language=\csname l@#1\endcsname
\fi
#2}}
\providecommand{\BIBdecl}{\relax}
\BIBdecl

\bibitem{WHO2016}
\BIBentryALTinterwordspacing
W.~H. Organization, ``Global report on diabetes,'' 2016. [Online]. Available:
  \url{http://apps.who.int/iris/bitstream/10665/204871/1/9789241565257\_eng.pdf}
\BIBentrySTDinterwordspacing

\bibitem{Farooq1}
M.~Farooq and E.~Sazonov, ``Accelerometer-based detection of food intake in
  free-living individuals,'' \emph{IEEE Sensors Journal}, vol.~18, no.~9, pp.
  3752--3758, May 2018.

\bibitem{Sazonov1}
E.~S. Sazonov and J.~M. Fontana, ``A sensor system for automatic detection of
  food intake through non-invasive monitoring of chewing,'' \emph{IEEE Sensors
  Journal}, vol.~12, no.~5, pp. 1340--1348, May 2012.

\bibitem{MyFitnessPal}
\BIBentryALTinterwordspacing
MyFitnessPal, ``Myfitnesspal, inc.'' 2018. [Online]. Available:
  \url{https://www.myfitnesspal.com/}
\BIBentrySTDinterwordspacing

\bibitem{FitNow}
\BIBentryALTinterwordspacing
FitNow, ``Fitnow, inc.'' 2018. [Online]. Available:
  \url{https://www.loseit.com/}
\BIBentrySTDinterwordspacing

\bibitem{Schoeller1}
D.~Schoeller, L.~Bandini, and W.~Dietz, ``Inaccuracies in self reported intake
  identified by comparison with the doubly labelled water method,'' \emph{Can.
  J. Physiol. Pharm.}, 1990.

\bibitem{Zhang4}
B.~S. A.~D. W.~Zhang, Q.~Yu and H.~Sawhney, ``Snap-n-eat: Food recognition and
  nutrition estimation on a smartphone,'' \emph{J. Diabetes Science and
  Technology}, vol.~9, no.~3, pp. 525--533, 2015.

\bibitem{Bi1}
Y.~Bi, M.~Lv, C.~Song, W.~Xu, N.~Guan, and W.~Yi, ``Autodietary: A wearable
  acoustic sensor system for food intake recognition in daily life,''
  \emph{IEEE Sensors Journal}, vol.~16, no.~3, pp. 806--816, 2016.

\bibitem{Mandal12}
B.~Mandal, L.~Li, V.~Chandrasekhar, and J.~H. Lim, ``Whole space subclass
  discriminant analysis for face recognition,'' in \emph{IEEE International Conference on Image Processing (ICIP)}, Quebec
  city, Canada, Sep 2015, pp. 329--333.

\bibitem{Mandal6}
B.~Mandal and H.-L. Eng, ``Regularized discriminant analysis for holistic human
  activity recognition,'' \emph{IEEE Intelligent Systems}, vol.~27, no.~1, pp.
  21--31, 2012.

\bibitem{Mandal14}
B.~Mandal, Z.~Wang, L.~Li, and A.~A. Kassim, ``Performance evaluation of local
  descriptors and distance measures on benchmarks and first-person-view videos
  for face identification,'' \emph{Neurocomputing}, vol. 184, pp. 107--116,
  2016.

\bibitem{Martinel1}
N.~Martinel, C.~Piciarelli, and C.~Micheloni, ``A supervised extreme learning
  committee for food recognition,'' \emph{Computer Vision and Image
  Understanding}, vol. 148, pp. 67--86, 2016.

\bibitem{Matsuda1}
M.~Yuji and K.~Yanai, ``Multiple-food recognition considering co-occurrence
  employing manifold ranking,'' in \emph{Proceedings of the $21^{st}$
  International Conference on Pattern Recognition}, Nov 2012.

\bibitem{Yanai1}
K.~Yanai and Y.~Kawano, ``Food image recognition using deep convolutional
  network with pre-training and fine-tuning,'' in \emph{2015 IEEE International
  Conference on Multimedia Expo Workshops (ICMEW)}, June 2015, pp. 1--6.

\bibitem{Chen5}
J.~Chen and C.-w. Ngo, ``Deep-based ingredient recognition for cooking recipe
  retrieval,'' in \emph{Proceedings of the ACM on Multimedia Conference}, 2016,
  pp. 32--41.

\bibitem{Shulin1}
S.~Yang, M.~Chen, D.~Pomerleau, and R.~Sukthankar, ``Food recognition using
  statistics of pairwise local features,'' in \emph{Proceedings of IEEE
  Conference on Computer Vision and Pattern Recognition}, Jun 2010.

\bibitem{Krizhevsky1}
A.~Krizhevsky, I.~Sutskever, and G.~E. Hinton, ``Imagenet classification with
  deep convolutional neural networks,'' in \emph{Advances in Neural Information
  Processing Systems}, 2012.

\bibitem{Liu14}
W.~Liu, Z.~Wang, and X.~L. \textit{et al.}, ``A survey of deep neural network
  architectures and their applications,'' \emph{Neurocomputing}, vol. 234, pp.
  11--26, 2017.

\bibitem{Pandey1}
P.~Pandey, A.~Deepthi, B.~Mandal, and N.~B. Puhan, ``Food\textsc{N}et:
  Recognizing foods using ensemble of deep networks,'' \emph{{IEEE} Signal
  Processing Letters}, vol.~24, no.~12, pp. 1758--1762, 2017.

\bibitem{Salehinejad1}
H.~Salehinejad, J.~Baarbe, S.~Sankar, J.~Barfett, E.~Colak, and S.~Valaee,
  ``Recent advances in recurrent neural networks,'' \emph{CoRR}, vol.
  abs/1801.01078, 2018.

\bibitem{Goodfellow1}
I.~Goodfellow, J.~Pouget-Abadie, M.~Mirza, B.~Xu, D.~Warde-Farley, S.~Ozair,
  A.~Courville, and Y.~Bengio, ``Generative adversarial nets,'' in
  \emph{Advances in Neural Information Processing Systems 27}, 2014, pp.
  2672--2680.

\bibitem{LeCun1}
Y.~LeCun, L.~Bottou, Y.~Bengio, and P.~Haffner, ``Gradient-based learning
  applied to document recognition,'' \emph{Proceedings of the IEEE}, vol.~86,
  no.~11, pp. 2278--2324, Nov 1998.

\bibitem{Krizhevsky2}
\BIBentryALTinterwordspacing
A.~Krizhevsky, ``Learning multiple layers of features from tiny images,'' 2009.
  [Online]. Available:
  \url{https://www.cs.toronto.edu/~kriz/learning-features-2009-TR.pdf}
\BIBentrySTDinterwordspacing

\bibitem{Denton1}
E.~L. Denton, S.~Chintala, A.~Szlam, and R.~Fergus, ``Deep generative image
  models using a laplacian pyramid of adversarial networks,'' in \emph{Advances
  in Neural Information Processing Systems 28: Annual Conference on Neural
  Information Processing Systems Montreal, Quebec, Canada}, 2015, pp.
  1486--1494.

\bibitem{Radford1}
\BIBentryALTinterwordspacing
A.~Radford, L.~Metz, and S.~Chintala, ``Unsupervised representation learning
  with deep convolutional generative adversarial networks,'' \emph{CoRR}, vol.
  abs/1511.06434, 2015. [Online]. Available:
  \url{http://arxiv.org/abs/1511.06434}
\BIBentrySTDinterwordspacing

\bibitem{Bossard1}
L.~Bossard, M.~Guillaumin, and L.~Van~Gool, ``Food-101 -- mining discriminative
  components with random forests,'' in \emph{European Conference on Computer
  Vision}, 2014, pp. 446--461.

\bibitem{SalimansGZCRC16}
\BIBentryALTinterwordspacing
T.~Salimans, I.~J. Goodfellow, W.~Zaremba, V.~Cheung, A.~Radford, and X.~Chen,
  ``Improved techniques for training gans,'' \emph{CoRR}, vol. abs/1606.03498,
  2016. [Online]. Available: \url{http://arxiv.org/abs/1606.03498}
\BIBentrySTDinterwordspacing

\bibitem{Christian1}
C.~Szegedy and \emph{et al.}, ``Going deeper with convolutions,'' in \emph{IEEE
  Conference on Computer Vision and Pattern Recognition (CVPR)}, 2015.

\bibitem{He4}
Z.~X. R. S.~e. He, K., ``Deep residual learning for image recognition,'' in
  \emph{{IEEE} {CVPR}}.\hskip 1em plus 0.5em minus 0.4em\relax IEEE, 2016, pp.
  770--778.

\bibitem{Kawano1}
Y.~Kawano and K.~Yanai, ``Real-time mobile food recognition system,'' in
  \emph{Computer Vision and Pattern Recognition Workshops (CVPRW)}, Jun 2013.

\end{thebibliography}

\end{document}